\documentclass[10pt,twocolumn,letterpaper]{article}

\usepackage{wacv}
\usepackage{times}
\usepackage{epsfig}
\usepackage{amsmath}
\usepackage{authblk}
\usepackage{amssymb}
\usepackage{graphicx}
\usepackage{subcaption}
\usepackage{url}
\usepackage{xcolor}
\usepackage{bibentry}
\usepackage[T1]{fontenc}

\usepackage{amsmath}
\renewcommand{\vec}[1]{\mathbf{#1}}

\definecolor{darkgreen}{rgb}{0.0, 0.5, 0}

\def\wacvPaperID{1171} 

\wacvfinalcopy 

\ifwacvfinal
\fi

\ifwacvfinal
\usepackage[breaklinks=true,bookmarks=false]{hyperref}
\else
\usepackage[pagebackref=true,breaklinks=true,colorlinks,bookmarks=false]{hyperref}
\fi

\ifwacvfinal
\else
\fi

\begin{document}

\title{Kernel Self-Attention for Weakly-supervised Image Classification using Deep Multiple Instance Learning}





\author[1,2,*]{
Dawid Rymarczyk
}
\author[1, 2,*]{Adriana Borowa}
\author[1,**]{Jacek Tabor}
\author[1, 2,**]{Bartosz Zieli\'nski}
\affil[1]{Faculty of Mathematics and Computer Science, Jagiellonian University\\
6~\L{}ojasiewicza Street, 30-348 Krak\'ow, Poland\\
}
\affil[2]{Ardigen SA\\76~Podole Street, 30-394 Krak\'ow, Poland\\
}
\affil[*]{\tt\small \{dawid.rymarczyk,ada.borowa\}@student.uj.edu.pl
}
\affil[**]{\tt\small \{jacek.tabor,bartosz.zielinski\}@uj.edu.pl
}

\maketitle
%

\begin{abstract}
Not all supervised learning problems are described by a pair of a fixed-size input tensor and a label. In some cases, especially in medical image analysis, a label corresponds to a bag of instances (e.g. image patches), and to classify such bag, aggregation of information from all of the instances is needed. There have been several attempts to create a model working with a bag of instances, however, they are assuming that there are no dependencies within the bag and the label is connected to at least one instance. In this work, we introduce Self-Attention Attention-based MIL Pooling (SA-AbMILP) aggregation operation to account for the dependencies between instances. We conduct several experiments on MNIST, histological, microbiological, and retinal databases to show that SA-AbMILP performs better than other models. Additionally, we investigate kernel variations of Self-Attention and their influence on the results.

\end{abstract}

\section{Introduction}

Classification methods typically assume that there exists a separate label for each example from a dataset. However, in many real-life applications, there exists only one label for a bag of instances because it is too laborious to label all of them separately. This type of problem, called Multiple Instance Learning (MIL)~\cite{dietterich1997solving}, assumes that there is only one label provided for the entire bag and that some of the instances associate to this label~\cite{foulds2010review}.

MIL problems are common in medical image analysis due to the vast resolution of images and the weakly-labeled small datasets~\cite{carbonneau2018multiple,quellec2017multiple}. Among others, they appear in the whole slide-image classification of biopsies ~\cite{campanella2019clinical,campanella2018terabyte,yoshida2018automated}, classification of dementia based on brain MRI ~\cite{tong2014multiple}, or the diabetic retinopathy screening~\cite{quellec2012multiple, rani2016multiple}. They are also used in computer-aided drug design to identify which conformers are responsible for the molecule activity~\cite{straehle2014multiple,zhao2013drug}.

Recently, Ilse \etal~\cite{ilse2018attention} introduced the Attention-based MIL Pooling (AbMILP), a trainable operator that aggregates information from multiple instances of a bag. It is based on a two-layered neural network with the attention weights, which allows finding essential instances. Since the publication, this mechanism was widely adopted in the medical image analysis~\cite{li2019attention,lu2019semi,wang2019rmdl}, especially for the assessment of whole-slide images. 
However, the Attention-based MIL Pooling is significantly different from the Self-Attention (SA) mechanism~\cite{zhang2018self}. It perfectly aggregates information from a varying number of instances, but it does not model dependencies between them. 
Additionally, the SA-AbMILP is distinct from other MIL approaches developed recently because it is not modeling the bag as a graph like in ~\cite{tu2019multiple, wang2018videos}, it is not using the pre-computed image descriptors as features as in ~\cite{tu2019multiple}, and it models the dependencies between instances in contrast to \cite{feng2017deep}.  

In this work, we introduce a method that combines self-attention with Attention-based MIL Pooling. It simultaneously catches the global dependencies between the instances in the bag (which are beneficial~\cite{zhou2009multi}) and aggregates them into a fixed-sized vector required for the successive layers of the network, which then can be used in regression, binary, and multi-class classification problems. Moreover, we investigate a broad spectrum of kernels replacing dot product when generating an attention map. According to the experiments' results, using our method with various kernels is beneficial compared to the baseline approach, especially in the case of more challenging MIL assumptions. Our code is publicly available at \url{https://github.com/gmum/Kernel_SA-AbMILP}.

\section{Multiple Instance Learning}
Multiple Instance Learning (MIL) is a variant of inductive machine learning belonging to the supervised learning paradigm~\cite{foulds2010review}. In a typical supervised problem, a separate feature vector, e.g. ResNet representation after Global Max Pooling, exists for each sample: $\vec{x} = \vec{h}, \vec{h} \in \mathbb{R}^{L \times 1}$. In MIL, each example is represented by a bag of feature vectors of length $L$ called instances: $\vec{x} = \{\vec{h_i}\}_{i=1}^n, \vec{h_i} \in \mathbb{R}^{L \times 1}$, and the bag is of variable size $n$. Moreover, in \emph{standard MIL assumption}, label of the bag $\vec{y} \in \{0,1\}$, each instance $h_i$ has a hidden binary label $y_i \in \{0,1\}$, and the bag is positive if at least one of its instances is positive:
\begin{equation}
\label{Y}
\centering \vec{y} = \begin{cases}
0, & \text{ iff }\sum \limits_{i=1}^{n}y_i = 0, \\ 
1, & \text{ otherwise.} 
\end{cases}
\end{equation}

This standard assumption (considered by AbMILP) is stringent and hence does not fit to numerous real-world problems. As an example, let us consider the digestive track assessment using the NHI scoring system~\cite{li2019simplified}, where the score $2$ is assigned to a biopsy if the neutrophils infiltrate more than 50\% of crypts and there is no epithelium damage or ulceration. Such a task obviously requires more challenging types of MIL~\cite{foulds2010review}, which operate on many assumptions (below defined as concepts) and classes.

Let $\hat{C} \subseteq C$ be the set of required instance-level concepts, and let $p: X \times C \rightarrow K$ be the function that counts how often the concept $c \in C$ occurs in the bag $\vec{x} \in X$. Then, in \emph{presence-based assumption}, the bag is positive if each concept occurs at least once:
\begin{equation}\label{Y_presence}
\centering \vec{y} = \begin{cases}
1, & \text{ iff for each } c \in \hat{C} : p(\vec{x}, c) \geq 1,\\ 
0, & \text{ otherwise. }
\end{cases}
\end{equation}

In the case of \emph{threshold-based assumptions}, the bag is positive if concept $c_i \in C$ occurs at least $t_i \in \mathbb{N}$ times:
\begin{equation}\label{Y_threshold}
\centering \vec{y} = \begin{cases}
1, & \text{ iff for each } c_i \in \hat{C}: p(\vec{x}, c_i) \geq t_i,\\
0, & \text{ otherwise. }
\end{cases}
\end{equation}

In this paper, we introduce methods suitable not only for the standard assumption (like it was in the case of AbMILP) but also for presence-based and threshold-based assumptions.

\section{Methods}
\label{sec:methods}

\subsection{Attention-based Multiple Instance Learning Pooling}
Attention-based MIL Pooling (AbMILP)~\cite{ilse2018attention} is a type of weighted average pooling, where the neural network determines the weights of instances. More formally, if the bag $\vec{x} = \{\vec{h_i}\}_{i=1}^n, \vec{h_i} \in \mathbb{R}^{L \times 1}$, then the output of the operator is defined as:
\begin{equation}\label{AbMILP_z}
\mathbf{z} = \sum_{i=1}^{n}a_i\mathbf{h_i},
\textup{ where }
a_i = \frac{\exp\left(\mathbf{w}^Ttanh(\mathbf{Vh_i})\right)}{\sum_j^N\exp\left(\mathbf{w}^Ttanh(\mathbf{Vh_j})\right)},
\end{equation}
$\mathbf{w} \in \mathbb{R}^{M \times 1}$ and $\mathbf{V} \in \mathbb{R}^{M \times L}$ are trainable layers of neural networks, and the hyperbolic tangent prevents the exploding gradient. Moreover, the weights $a_i$ sum up to $1$ to wean from various sizes of the bags and the instances are in the random order within the bag to prevent the overfitting. 

The most important limitation of AbMILP is the assumption that all instances of the bag are independent. To overcome this limitation, we extend it by introducing the Self-Attention (SA) mechanism~\cite{zhang2018self} which models dependencies between instances of the bag.

\begin{figure*}[!t]
	\centering
	\includegraphics[width=0.99\linewidth]{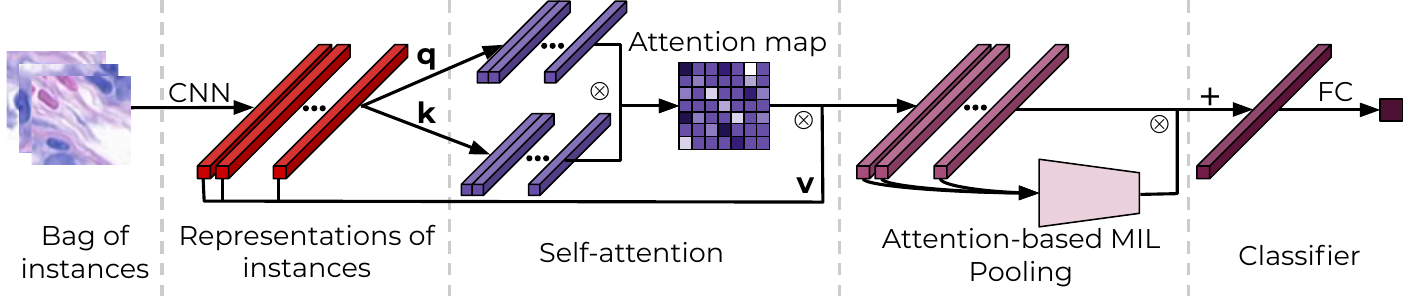} 
	\caption{The pipeline of self-attention in deep MIL starts with obtaining feature space representation for each of the instances from the bag using features block of Convolutional Neural Network (CNN). In order to model dependencies between the instances, their representations pass trough the self-attention layer and then aggregate using AbMILP operator. The obtained fixed-size vector goes trough the Fully Connected (FC) classification layer.}
	\label{fig:MIL_SA}
\end{figure*}

\subsection{Self-Attention in Multiple Instance Learning}
\label{subsec:SAiMIL}
The pipeline of our method, which applies Self-Attention into Attention-based MIL Pooling (SA-AbMILP), consists of four steps. First, the bag's images are passed through the Convolutional Neural Network (CNN) to obtain their representations. Those representations are used by the self-attention module (with dot product or the other kernels) to integrate dependencies of the instances into the process. Feature vectors with integrated dependencies are used as the input for the AbMILP module to obtain one fixed-sized vector for each bag. Such a vector can be passed to successive Fully-Connected (FC) layers of the network. The whole pipeline is presented in Fig.~\ref{fig:MIL_SA}. In order to make this work self-contained, below, we describe self-attention and particular kernels.

\paragraph{Self-Attention (SA).} SA is responsible for finding the dependencies between instances within one bag. Instance representation after SA is enriched with the knowledge from the entire bag, this is important for the detection of the number of instances of the same concept and their relation. SA transforms all the instances into two feature spaces of keys $\mathbf{k_i} = \mathbf{W_kh_i}$ and queries $\mathbf{q_j} = \mathbf{W_qh_j}$, and calculates:
\begin{equation}\label{att_B}
\mathbf{\beta_{j,i}} = \frac{\exp{(s_{ij})}}{\sum_{i=1}^N\exp{(s_{ij})}},
\textup{ where }
s_{ij} = \langle\mathbf{k}(\mathbf{h_i}),\mathbf{q}(\mathbf{h_j})\rangle,
\end{equation}
to indicate the extent to which the model attends to the $i^{th}$ instance when synthesizing the $j^{th}$ one. The output of the attention layer is defined separately for each instance as:
\begin{equation}\label{att_o}
\mathbf{\hat{h}_{j}} = \gamma\mathbf{o_j} + \mathbf{h_j},
\textup{ where }
\mathbf{o_{j}} = \sum_{i=1}^{N}\beta_{j, i}\mathbf{W_v}\mathbf{h_i},
\end{equation}
$\mathbf{W_q}, \mathbf{W_k} \in \mathbb{R}^{\bar{L} \times L}$, $\mathbf{W_v} \in \mathbb{R}^{L \times L}$ are trainable layers, $\bar{L} = L / 8$, and $\gamma$ is a trainable scalar initialized to $0$. Parameters $\bar{L}$ and $\gamma$ were chosen based on the results presented in \cite{zhang2018self}.

\paragraph{Kernels in self-attention.} In order to indicate to which extent one instance attends on synthesizing the other one, the self-attention mechanism typically employs a dot product (see $s_{ij}$ in Eq.~\ref{att_B}). However, dot product can be replaced by a various kernel with positive results observed in Support Vectors Machine (SVM)~\cite{baudat2001kernel} or Convolutional Neural Networks (CNN)~\cite{wang2019kervolutional}, especially in the case of small training sets.

The Radial Basis Function (RBF) and Laplace kernels were already successfully adopted to self-attention~\cite{kim2019attentive,tsai2019transformer}. Hence, in this study, we additionally extend our approach with the following standard choice of kernels (with $\alpha$ as a trainable parameter):
\begin{itemize}
    \item Radial Basis Function (GSA-AbMILP): \\$k(x, y) = \exp{(-\alpha \|x - y\|_2^2)}$,
    \item Inverse quadratic (IQSA-AbMILP): \\$k(x, y) = \frac{1}{\alpha\|x - y\|_2^2 + 1}$,
    \item Laplace (LSA-AbMILP): \\$k(x,y)=-\|x-y\|_1$,
    \item Module (MSA-AbMILP):  \\$k(x, y) = \|x-y\|^\alpha-\|x\|^\alpha-\|y\|^\alpha$.
\end{itemize}

\noindent We decided to limit to those kernels because they are complementary regarding the shape of tails in their distributions.

\section{Experiments}
We adopt five datasets, from which four were adopted to MIL algorithms~\cite{ilse2018attention, tu2019multiple},  to investigate the performance of our method: MNIST~\cite{lecun1998gradient}, two histological databases of colon~\cite{sirinukunwattana2016locality} and breast~\cite{gelasca2008evaluation} cancer, a microbiological dataset DIFaS~\cite{zielinski2019deep}, and a diabetic retinopathy screening data set called “Messidor”~\cite{decenciere2014feedback}. For MNIST, we adapt LeNet5~\cite{lecun1998gradient} architecture, for both histological datasets SC-CNN~\cite{sirinukunwattana2016locality} is applied as it was in~\cite{ilse2018attention}, for microbiological dataset we use convolutional parts of ResNet-18~\cite{he2016deep} or AlexNet~\cite{krizhevsky2012imagenet} followed by $1\times 1$ convolution as those were the best feature extractor in~\cite{zielinski2019deep}, and for the "Messidor" dataset we used the ResNet-18~\cite{he2016deep} as most of the approaches that we compare here are based on the handcrafted image features like in~\cite{tu2019multiple}. The experiments, for MNiST, histological and "Messidor" datasets, are repeated $5$ times using $10$ fold cross-validation with $1$ validation fold and $1$ test fold. In the case of the microbiological dataset, we use the original $2$ fold cross-validation which divides the images by the preparation, as using images from the same preparation in both training and test set can result in overstated accuracy~\cite{zielinski2019deep}. Due to the dataset complexity, we use the early stopping mechanism with different windows: $5$, $25$, $50$ and $70$ epochs for MNIST, histological datasets, microbiological, and "Messidor" datasets, respectively.
We compare the performance of our method (SA-AbMILP) and its kernel variations with instance-level approaches (instance+max, instance+mean, and instance+voting), embedding-level approaches (embedding+max and embedding+mean), and Attention-based MIL Pooling, AbMILP~\cite{ilse2018attention}. The instance-level approaches in the case of MNIST and histological database compute the maximum or mean value of the instance scores. For the microbiological database, instance scores are aggregated by voting due to multiclassification. The embedding-level approaches calculate the maximum or mean for feature vector of the instances. For "Messidor" dataset we are comparing to the results obtained by~\cite{tu2019multiple, feng2017deep}.
We run a Wilcoxon signed-rank test on the results to identify which ones significantly differ from each other, and which ones do not (and thus can be considered equally good). The comparison is performed between the best method (the one with the best mean score) and all the other methods, for each experiment separately. The mean accuracy is obtained as average over $5$ repetitions with the same train/test divisions used by all compared methods. The number of repetitions is relatively small for statistical tests. Therefore we set the p-value to $0.1$. For computations, we use Nvidia GeForce RTX 2080.

\begin{figure*}[!ht]
    \centering
    \includegraphics[width=\linewidth]{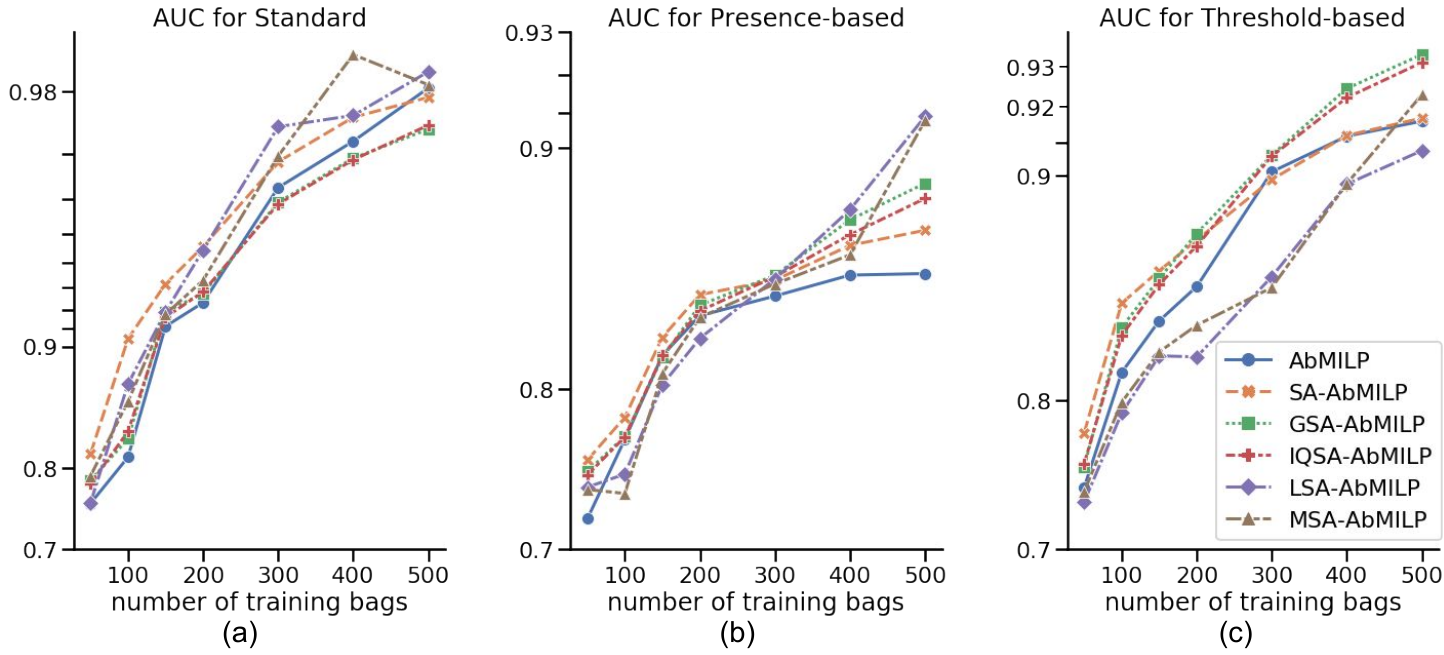}
    \caption{Results for MNIST dataset with bags generated using standard (a), presence-based (b), and threshold-based (c) assumption. In all cases, our approach, either with dot product (SA-AbMILP) or the other kernels (GSA-AbMILP, IQSA-AbMILP, LSA-AbMILP, and MSA-AbMILP) obtains statistically better results than the baseline method (AbMILP). See Section~\ref{sec:methods} for description of the shortcuts.}\label{fig:MNISST_BAGS}
\end{figure*}

\subsection{MNIST dataset}

\paragraph{Experiment details.} As in \cite{ilse2018attention}, we first construct various types of bags based on the MNIST dataset. Each bag contains a random number of MNIST images (drawn from Gaussian distributions $\mathcal{N}(10, 2)$). We adopt three types of bag labels referring to three types of MIL assumptions:
\begin{itemize}
    \item Standard assumptions: $\vec{y}=1$ if there is at least one occurrence of ``9'',
    \item Presence-based assumptions: $\vec{y}=1$ if there is at least one occurrence of ``9'' and at least one occurrence of ``7'',
    \item Threshold-based assumptions: $\vec{y}=1$ if there are at least two occurrences of ``9''.
\end{itemize}
We decided to use ``9'' and ``7'' because they are often confused with each other, making the task more challenging.

We investigated how the performance of the model depends on the number of bags used in training (we consider 50, 100, 150, 200, 300, 400, and 500 training bags). For all experiments, we use LeNet5~\cite{lecun1998gradient} initialized according to~\cite{glorot2010understanding} with the bias set to 0. We use Adam optimizer~\cite{kingma2014adam} with parameters $\beta_1=0.9$ and $\beta_2=0.999$, learning rate $10^{-5}$, and batch size 1.

\begin{figure*}
    \centering
    \includegraphics[width=0.8\linewidth]{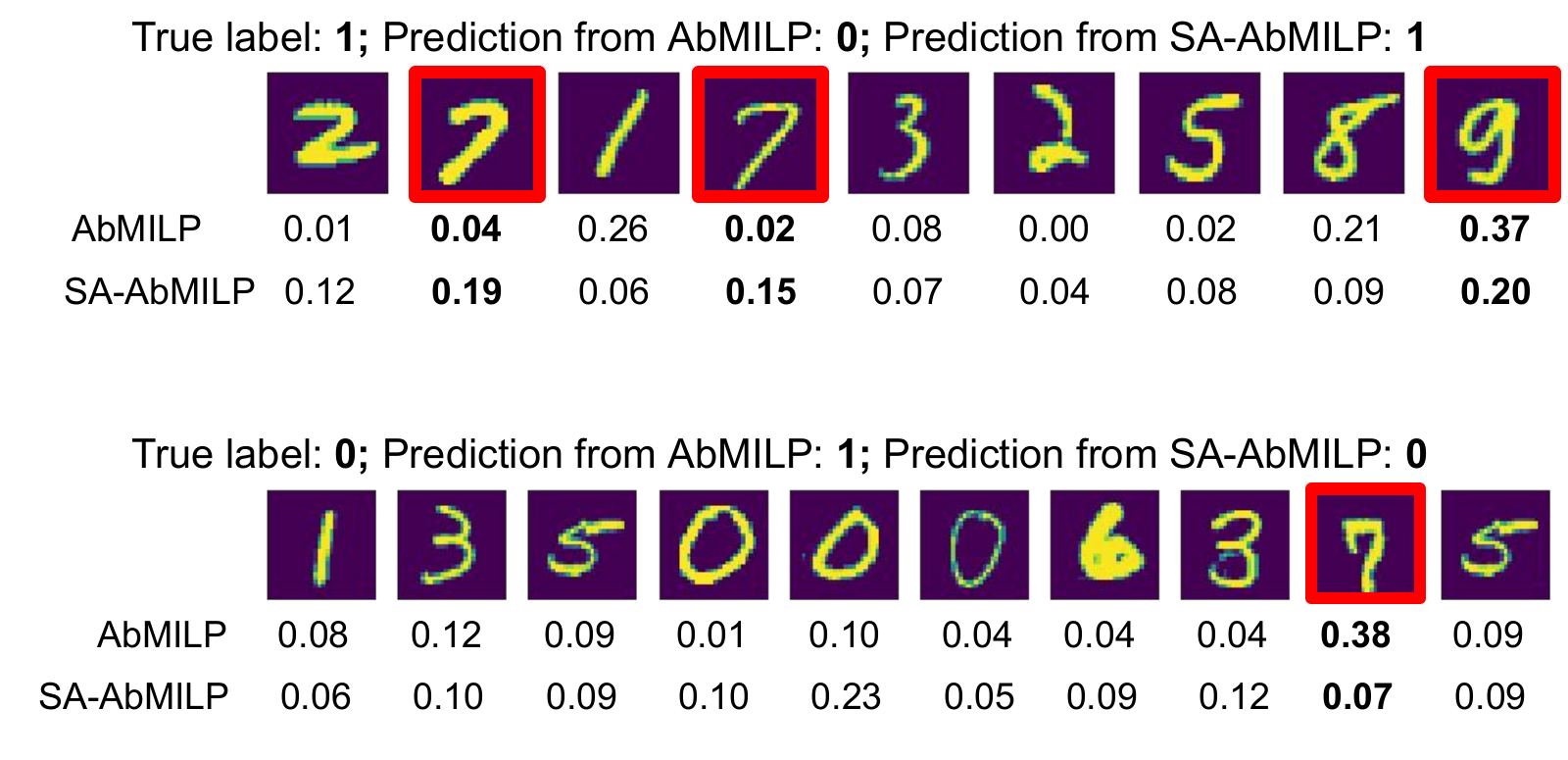}
    \caption{Example of instances' weights for a positive (top) and negative (bottom) bag in a presence-based assumption (where positive is at least one occurrence of ``9'' and ``7'') for AbMILP and our method. One can observe that for SA-AbMILP, ``9''s and ``7''s strengthen each other in the self-attention module, resulting in higher weights in the aggregation operator than for AbMILP.}
    \label{fig:PresenceWeights}
\end{figure*}

\paragraph{Results.} AUC values for considered MIL assumptions are visualized in Fig.~\ref{fig:MNISST_BAGS}. One can observe that our method with a dot product (SA-AbMILP) always outperforms other methods in case of small datasets. However, when the number of training examples reaches 300, its kernel extensions work better (Laplace in presence-based and inverse quadratic in threshold-based assumption). Hence, we conclude that for small datasets, no kernel extensions should be applied, while in the case of a larger dataset, a kernel should be optimized together with the other hyperparameters. Additionally, we analyze differences between the weights of instances in AbMILP and our method. As presented in Fig.~\ref{fig:PresenceWeights}, for our method, ``9''s and ``7''s strengthen each other in the self-attention module, resulting in higher weights in the aggregation operator than for AbMILP, which returns high weight for only one digit (either ``9'' or ``7'').

\subsection{Histological datasets}

\paragraph{Experiment details.} In the second experiment, we consider two histological datasets of \textit{breast} and \textit{colon} cancer (described below). For both of them, we generate instance representations using SC-CNN~\cite{sirinukunwattana2016locality} initialized according to~\cite{glorot2010understanding} with the bias set to 0. We use Adam~\cite{kingma2014adam} optimizer with parameters $\beta_1=0.9$ and $\beta_2=0.999$, learning rate $10^{-4}$, and batch size 1. We also apply extensive data augmentation, including random rotations, horizontal and vertical flipping, random staining augmentation~\cite{ilse2018attention}, staining normalization~\cite{tomczak2018histopathological}, and instance normalization.

\paragraph{Breast cancer dataset.} Dataset from~\cite{gelasca2008evaluation} contains $58$ weakly labeled H\&E biopsy images of resolution $896 \times 768$. The image is labeled as malignant if it contains at least one cancer cell. Otherwise, it is labeled as benign. Each image is divided into patches of resolution $32\times32$, resulting in $672$ patches per image. Patches with at least 75\% of the white pixels are discarded, generating $58$ bags of various sizes.


\paragraph{Colon cancer dataset.} Dataset from~\cite{sirinukunwattana2016locality} contains $100$ images with $22444$ nuclei manually assigned to one of the following classes: epithelial, inflammatory, fibroblast, and miscellaneous. We construct bags of $27\times27$ patches with centers located in the middle of the nuclei. The bag has a positive label if there is at least one epithelium nucleus in the bag. Tagging epithelium nuclei is essential in the case of colon cancer because the disease originates from them~\cite{ricci2007identification}.

\begin{center}
\begin{table*}[!ht]
\caption{Results for breast and colon cancer datasets (mean and standard error of the mean over $5$ repetitions). See Section~\ref{sec:methods} for description of the acronyms.}
\centering
 \begin{tabular}{||c | c | c | c | c | c ||} 
  \hline
 \multicolumn{6}{|c|}{breast cancer dataset} \\
 \hline
 method & accuracy & precision & recall & F-score & AUC \\ [0.5ex] 
 \hline\hline
 instance+max                   & $61.4 \pm 2.0$ & $58.5 \pm 3.0$ & $47.7 \pm 8.7$ & $50.6 \pm 5.4$ & $61.2 \pm 2.6$ \\ 
 \hline
 instance+mean                  & $67.2 \pm 2.6$ & $67.2 \pm 3.4$ & $51.5 \pm 5.6$ & $57.7 \pm 4.9$ & $71.9 \pm 1.9$ \\ 
 \hline
 embedding+max                  & $60.7 \pm 1.5$ & $55.8 \pm 1.3$ & $54.6 \pm 7.0$ & $54.3 \pm 4.2$ & $65.0 \pm 1.3$ \\ 
 \hline
 embedding+mean                 & $\mathbf{74.1 \pm 2.3}$ & $74.1 \pm 2.3$ & $65.4 \pm 5.4$ & $\mathbf{68.9 \pm 3.4}$ & $79.6 \pm 1.2$ \\ 
 \hline
 AbMILP                         & $71.7 \pm 2.7$ & $\mathbf{77.1 \pm 4.1}$ & $68.6 \pm 3.9$ & $66.5 \pm 3.1$ & $85.6 \pm 2.2$ \\ 
 \hline
 SA-AbMILP                      & $\mathbf{75.1 \pm 2.4}$ & $\mathbf{77.4 \pm 3.7}$ & $74.9 \pm 3.7$ & $\mathbf{69.9 \pm 3.0}$ & $\mathbf{86.2 \pm 2.2}$ \\
 \hline
GSA-AbMILP                     & $\mathbf{75.8 \pm 2.2}$ & $\mathbf{79.3 \pm 3.3}$ & $74.7 \pm 3.4$ & $\mathbf{72.5 \pm 2.5}$ & $\mathbf{85.9 \pm 2.2}$ \\
 \hline
 IQSA-AbMILP                    & $\mathbf{76.7 \pm 2.3}$ & $\mathbf{78.6 \pm 4.2}$ & $75.1 \pm 4.2$ & $\mathbf{66.6 \pm 4.3}$ & $\mathbf{85.9 \pm 2.2}$ \\
 \hline
 LSA-AbMILP                     & $65.5 \pm 2.9$ & $62.5 \pm 3.7$ & $\mathbf{89.5 \pm 2.6}$ & $\mathbf{68.5 \pm 2.6}$ & $\mathbf{86.7 \pm 2.1}$ \\
 \hline
 MSA-AbMILP                     & $\mathbf{73.8 \pm 2.6}$ & $\mathbf{78.4 \pm 3.9}$ & $73.8 \pm 3.6$ & $\mathbf{69.4 \pm 3.4}$ & $85.8 \pm 2.2$ \\ 
 \hline
 \hline
 \multicolumn{6}{|c|}{colon cancer dataset} \\
 \hline
 method & accuracy & precision & recall & F-score & AUC\\ [0.5ex] 
 \hline\hline
 instance+max & $84.2 \pm 2.1$ & $86.6 \pm 1.7$ & $81.6 \pm 3.1$ & $83.9 \pm 2.3$ & $91.4 \pm 1.0$ \\ 
 \hline
 instance+mean & $77.2 \pm 1.2$ & $82.1 \pm 1.1$ & $71.0 \pm 3.1$ & $75.9 \pm 1.7$ & $86.6 \pm 0.8$ \\ 
 \hline
 embedding+max & $82.4 \pm 1.5$ & $88.4 \pm 1.4$ & $75.3 \pm 2.0$ & $81.3 \pm 1.7$ & $91.8 \pm 1.0$ \\ 
 \hline
 embedding+mean & $86.0 \pm 1.4$ & $81.1 \pm 1.1$ & $80.4 \pm 2.7$ & $85.3 \pm 1.6$ & $94.0 \pm 1.0$ \\ 
 \hline
 AbMILP & $88.4 \pm 1.4$ & $\mathbf{95.3 \pm 1.5}$ & $\mathbf{84.1 \pm 2.9}$ & $\mathbf{87.2 \pm 2.1}$ & $97.3 \pm 0.7$ \\ 
 \hline
 SA-AbMILP & $\mathbf{90.8 \pm 1.3}$ & $\mathbf{93.8 \pm 2.0}$ & $\mathbf{87.2 \pm 2.4}$ & $\mathbf{89.0 \pm 1.9}$ & $\mathbf{98.1 \pm 0.7}$ \\
 \hline
 GSA-AbMILP & $88.4 \pm 1.7$ & $\mathbf{95.2 \pm 1.7}$ & $83.7 \pm 2.8$ & $\mathbf{86.9 \pm 2.1}$ & $\mathbf{98.5 \pm 0.6}$ \\
 \hline
 IQSA-AbMILP & $89.0 \pm 1.9$ & $\mathbf{93.9 \pm 2.1}$ & $\mathbf{85.5 \pm 3.0}$ & $\mathbf{86.9 \pm 2.5}$ & $96.6 \pm 1.1$ \\
 \hline
 LSA-AbMILP & $84.8 \pm 1.8$ & $\mathbf{92.7 \pm 2.7}$ & $71.1 \pm 4.6$ & $73.4 \pm 4.3$ & $95.5 \pm 1.7$ \\ 
 \hline
 MSA-AbMILP & $\mathbf{89.6 \pm 1.6} $ & $\mathbf{94.6 \pm 1.5}$ & $\mathbf{85.7 \pm 2.7}$ & $\mathbf{87.9 \pm 1.8}$ & $\mathbf{98.4 \pm 0.5}$ \\
 \hline
 \end{tabular}
 \label{tab:cancer_results}
 \end{table*}
 \end{center}

\begin{center}
\begin{table*}[!ht]
\caption{Results for DIFaS dataset (mean and standard error of the mean over $5$ repetitions). See Section~\ref{sec:methods} for description of the shortcuts.}
\centering
 \begin{tabular}{||c | c | c | c | c | c ||} 
  \hline
 \multicolumn{6}{|c|}{DIFaS (ResNet-18)} \\
 \hline
 method & accuracy & precision & recall & F-score & AUC \\ [0.5ex] 
 \hline\hline
 instance+voting                  & $78.3 \pm 2.0$ & $78.0 \pm 1.4$ & $76.0 \pm 1.7$ & $75.8 \pm 2.0$ & $N/A$ \\ 
 \hline
 embedding+max                  & $77.1 \pm 0.7$ & $\mathbf{83.1 \pm 0.5}$ & $77.1 \pm 0.7$ & $75.5 \pm 0.9$ & $95.3 \pm 0.2$ \\ 
 \hline
 embedding+mean                 & $78.1 \pm 0.8$ & $\mathbf{83.3 \pm 0.5}$ & $78.1 \pm 0.8$ & $76.4 \pm 1.0 $ & $95.2 \pm 0.2$ \\ 
 \hline
 AbMILP                         & $77.5 \pm 0.6$ & $82.6 \pm 0.5$ & $77.5 \pm 0.6$ & $75.6 \pm 0.8$ & $96.1 \pm 0.3$ \\ 
 \hline
 SA-AbMILP                      & $\mathbf{80.1 \pm 0.6}$ & $\mathbf{84.6 \pm 0.6}$ & $\mathbf{80.1 \pm 0.6}$ & $\mathbf{78.4 \pm 0.8}$ & $\mathbf{96.8 \pm 0.3}$ \\
 \hline
GSA-AbMILP                     & $\mathbf{79.1 \pm 0.4}$ & $83.5 \pm 0.7$ & $\mathbf{79.1 \pm 0.4}$ & $77.2 \pm 0.5$ & $\mathbf{97.0 \pm 0.3}$ \\
 \hline
 IQSA-AbMILP                    & $\mathbf{79.4 \pm 0.4}$ & $\mathbf{83.7 \pm 0.7}$ & $\mathbf{79.4 \pm 0.4}$ & $\mathbf{77.6 \pm 0.6}$ & $96.8 \pm 0.3$ \\
 \hline
 LSA-AbMILP                     & $77.6 \pm 0.5$ & $82.5 \pm 0.5$ & $77.6 \pm0.5$ & $75.7 \pm 0.7$ & $96.2 \pm 0.3$ \\
 \hline
 MSA-AbMILP                     & $\mathbf{79.2 \pm 0.4}$ & $83.5 \pm 0.4$ & $\mathbf{79.2 \pm 0.4}$ & $\mathbf{77.5 \pm 0.6}$ & $\mathbf{96.9 \pm 0.3}$ \\ 
 \hline
 \hline
 \multicolumn{6}{|c|}{DIFaS (AlexNet)} \\
 \hline
 method & accuracy & precision & recall & F-score & AUC\\ [0.5ex] 
 \hline\hline
 instance+voting                  & $77.3 \pm 1.9$ & $78.4 \pm 0.8$ & $76.6 \pm 1.2$ & $76.2 \pm 1.0$ & $N/A$ \\ 
 \hline
 embedding+max & $82.9 \pm 1.2$ & $87.1 \pm 0.9$ & $82.9 \pm 1.2$ & $82.3 \pm 1.4$ & $98.4 \pm 0.2$ \\ 
 \hline
 embedding+mean & $82.3 \pm 0.7$ & $87.2 \pm 0.4$ & $82.3 \pm 0.7$ & $81.5 \pm 1.0$ & $98.1 \pm 0.3$ \\ 
 \hline
 AbMILP & $83.6 \pm 1.2$ & $87.8 \pm 0.7$ & $83.6 \pm 1.2$ & $82.9 \pm 1.5$ & $98.6 \pm 0.2$ \\ 
 \hline
 SA-AbMILP & $\mathbf{86.0 \pm 1.0}$ & $\mathbf{89.6 \pm 0.6}$ & $\mathbf{86.0 \pm 1.0}$ & $\mathbf{85.7 \pm 1.3}$ & $\mathbf{98.9 \pm 0.1}$ \\
 \hline
 GSA-AbMILP & $84.6 \pm 1.0$ & $89.1 \pm 0.6$ & $84.6 \pm 1.0$ & $84.2 \pm 1.3$ & $98.8 \pm 0.2$ \\
 \hline
 IQSA-AbMILP & $84.1 \pm 1.2$ & $88.4 \pm 0.6$ & $84.1 \pm 1.2$ & $83.4 \pm 1.4$ & $\mathbf{98.9 \pm 0.2}$ \\
 \hline
 LSA-AbMILP & $83.9 \pm 1.3$ & $88.0 \pm 0.7$ & $83.9 \pm 1.3$ & $83.2 \pm 1.6$ & $98.6 \pm 0.2$ \\ 
 \hline
 MSA-AbMILP & $83.1 \pm 0.8  $ & $87.8 \pm 0.4$ & $83.1 \pm 0.8$ & $82.4 \pm 1.0$ & $98.7 \pm 0.2$ \\
 \hline
 \end{tabular}
 \label{tab:fungus_results}
 \end{table*}
 \end{center}
 
 \begin{center}
 \begin{table}
 \centering
 \caption{Retinal image screening dataset results. *Results with asterisk are sourced from \cite{tu2019multiple}.}
 \begin{tabular}{||c | c | c ||} 
  \hline
 \multicolumn{3}{|c|}{Retinal image screening dataset} \\
 \hline
 method & accuracy & F-score \\ [0.5ex] 
 \hline\hline
LSA-AbMILP & $\mathbf{76.3}\%$ & $\mathbf{0.77}$ \\
\hline
SA-AbMILP & $75.2\%$ & $0.76$ \\
\hline
AbMILP & $74.5\%$ & $0.74$ \\
\hline
GSA-AbMILP & $74.5\%$ & $0.75$ \\
\hline
IQSA-AbMILP & $74.5\%$ & $0.75$ \\
\hline
MSA-AbMILP & $73.5\%$ & $0.74$ \\
\hline
\hline
MIL-GNN-DP*   & $74.2\%$ & $\mathbf{0.77}$ \\ 
\hline
MIL-GNN-Att* & $72.9\%$ & $0.75$ \\
\hline
mi-Graph* & $72.5\%$ & $0.75$ \\
\hline
MILBoost* & $64.1\%$ & $0.66$ \\
\hline
Citation k-NN* & $62.8\% $ & $0.68$ \\
\hline
EMDD* & $55.1\%$ & $0.69$ \\
\hline
MI-SVM* & $54.5\%$ & $0.70$ \\
\hline
mi-SVM* & $54.5\%$ & $0.71$ \\
 \hline
 \end{tabular}
 \label{tab:retina_results}

 \end{table}
 \end{center}

\begin{figure*}[!ht]
    \centering
    \includegraphics[width=\linewidth]{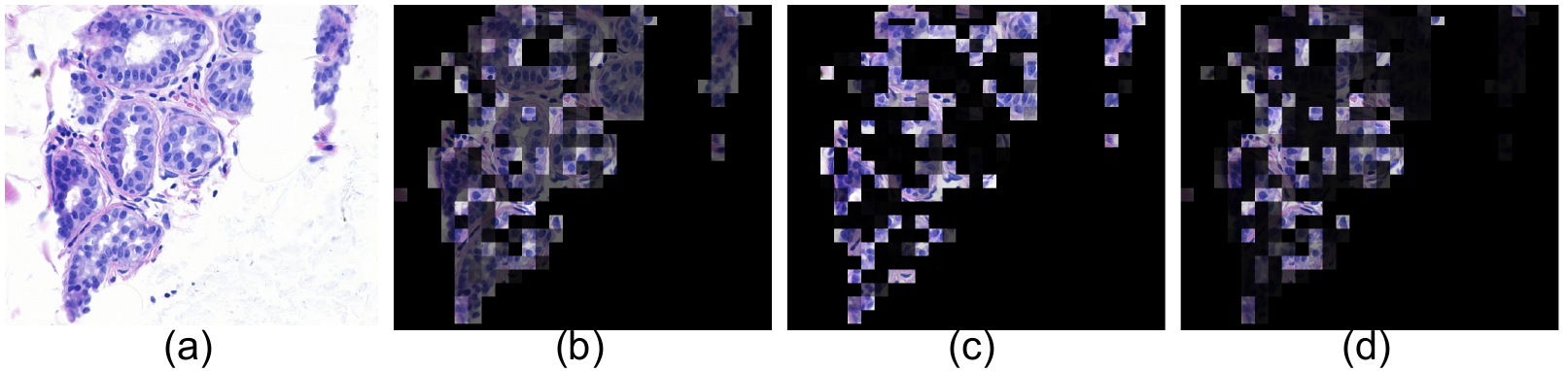}
    \caption{An example image from the breast cancer dataset~(a), weights of patches obtained by AbMILP~(b) and SA-AbMILP~(c), and SA dependencies obtained for the most relevant patch in SA-AbMILP~(d). One can observe that SA-AbMILP highlights fewer patches than AbMILP, which simplifies their analysis. Additionally, SA dependencies are justified histologically, as they mostly focus on nuclei located in the neighborhood of crypts.}
    \label{fig:breast_diff}
\end{figure*}

\paragraph{Results.} Results for histological datasets are presented in Table~\ref{tab:cancer_results}. For both of them, our method (with or without kernel extension) improves the Area Under the ROC Curve (AUC) comparing to the baseline methods. Moreover, our method obtains the highest recall, which is of importance for reducing the number of false negatives. To explain why our method surpasses the AbMILP, we compare the weights of patches in the average pooling. Those patches contribute the most to the final score and should be investigated by the pathologists. One can observe in Fig.~\ref{fig:breast_diff} that our method highlights fewer patches than AbMILP, which simplifies their analysis. Additionally, SA dependencies obtained for the most relevant patch of our method are justified histologically, as they mostly focus on nuclei located in the neighborhood of crypts. Moreover, in the case of the colon cancer dataset, we further observe the positive aspect of our method, as it strengthens epithelium nuclei and weakens nuclei in the lamina propria at the same time,. Finally, we notice that kernels often improve overall performance but none of them is significantly superior.

\begin{figure*}[!ht]
    \centering
    \includegraphics[width=\linewidth]{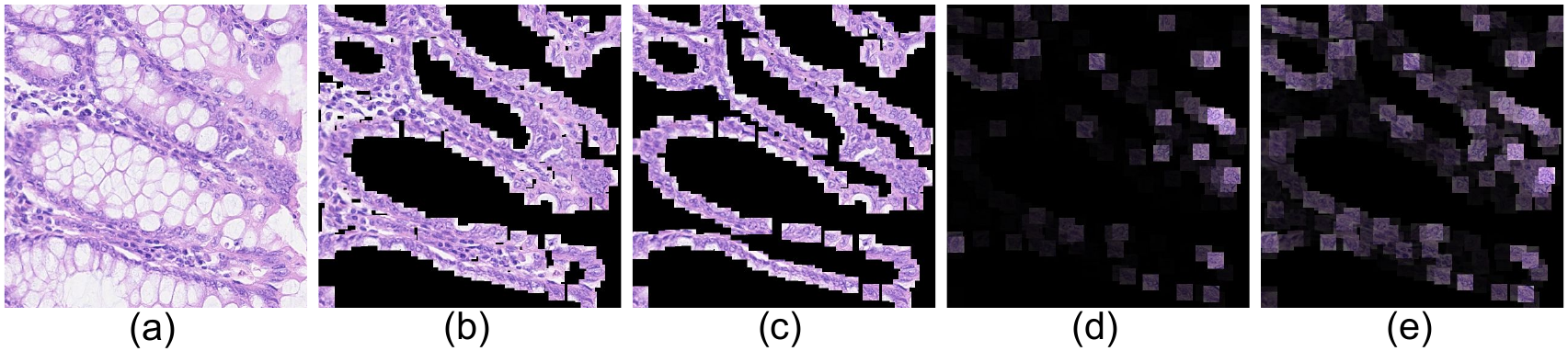}
    \caption{An example image from the colon cancer dataset~(a) with annotated nuclei~(b) and epithelium nuclei~(c), as well as, the weights of patches obtained by AbMILP~(d) and SA-AbMILP~(e). One can observe that SA-AbMILP strengthens epithelium nuclei and, at the same time, weakens nuclei in the lamina propria.}
    \label{fig:colon_diff}
\end{figure*}

\begin{figure*}[!ht]
    \centering
    \includegraphics[width=\linewidth]{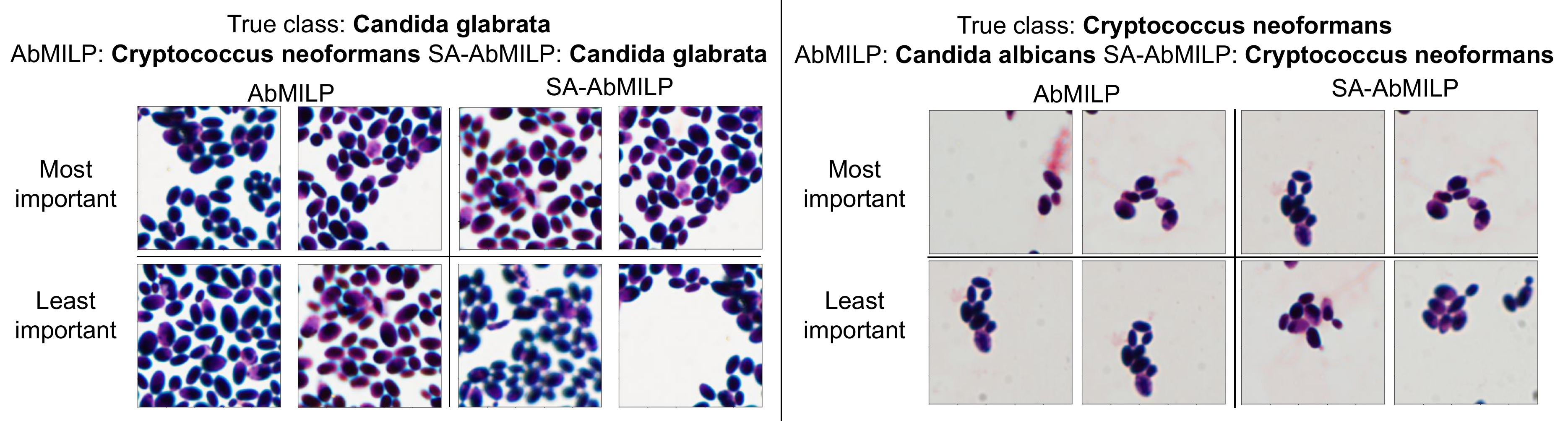}
    \caption{Example patches from the DIFaS dataset with the lowest and highest weights in the average pooling. One can observe that SA-AbMILP properly~\cite{zielinski2019deep} assigns lower weights to blurred patches with a small number of cells. Moreover, in contrast to the AbMILP, it assigns high weight to clean patches (without red artifacts).}
    \label{fig:fungus_patches}
\end{figure*}

\subsection{Microbiological dataset}

\label{sect:fungus}

\paragraph{Experiment details.} In the final experiment, we consider the microbiological DIFaS database~\cite{zielinski2019deep} of \textit{fungi species}. It contains $180$ images for $9$ fungi species (there are $2$ preparations with $10$ images for each species and it is a multi-class classification). As the size of images is $5760\times 3600\times 3$ pixels, it is difficult to process the entire image through the network so we generate patches following method used in~\cite{zielinski2019deep}. Hence, unlike the pipeline from Section~\ref{subsec:SAiMIL}, we use two separate networks. The first network generates the representations of the instances, while the second network (consisting of self-attention, attention-based MIL pooling, and classifier) uses those representations to recognize fungi species. Due to the separation, it is possible to use deep architectures like ResNet-18~\cite{he2016deep} and AlexNet~\cite{krizhevsky2012imagenet} pre-trained on ImageNet database~\cite{deng2009imagenet} to generate the representations. The second network is trained using Adam~\cite{kingma2014adam} optimizer with parameters $\beta_1=0.9$ and $\beta_2=0.999$, learning rate $10^{-5}$, and batch size $1$. We also apply extensive data augmentation, including random rotations, horizontal and vertical flipping, Gaussian noise, and patch normalization. To increase the training set, each iteration randomly selects a different subset of image's patches as an input.

\paragraph{Results.} Results for DIFaS database are presented in Table~\ref{tab:fungus_results}. Our method improves almost all of the scores for both feature pooling networks. The exception is the precision for ResNet-18, where our method is on par with maximum and mean instance representation. Moreover, we observe that while non-standard kernels can improve results for representations obtained with ResNet-18, they do not operate well with those generated with AlexNet. Additionally, to interpret the model outputs, we visualize patches with the lowest and highest weights in the average pooling. As shown in Figure~\ref{fig:fungus_patches}, our method properly~\cite{zielinski2019deep} assigns lower weights to blurred patches with a small number of cells. Also, in contrast to the baseline method, it assigns high weight to clean patches (without red artifacts).

\subsection{Retinal image screening dataset.} 

\paragraph{Experiment details.} Dataset "Messidor" from~\cite{decenciere2014feedback} contains $1200$ images with $654$ positive (diagnosed with diabetes) and $546$ negative (healthy) images. The size of each image is $700\times700$ pixels. Each image is partitioned into patches of $224\times224$ pixels. Patches containing only background are dropped. We are using ResNet18~\cite{he2016deep} pretrained on the ImageNet~\cite{deng2009imagenet} as an instance feature vectors generator and SA-AbMILP to obtain the final prediction, which is trained in an end to end fashion. The model is trained using Adam~\cite{kingma2014adam} optimizer with parameters $\beta_1=0.9$ and $\beta_2=0.999$, learning rate $5*10^{-6}$, and batch size $1$. We also apply data augmentation as in Section~\ref{sect:fungus}.

\paragraph{Results.} Results for "Messidor" database are presented in Table~\ref{tab:retina_results} alongside results of other approaches which are used for comparison and were taken from~\cite{tu2019multiple}. Our method improves accuracy and the highest scores are achieved using Laplace kernel variation, which obtains F1 score on par with the best reference approach. This is the only database for which Laplace kernel obtains the best results, which only confirms that the kernel should be optimized together with the other hyperparameters when applied to new problems.

\section{Conclusions and discussion}
In this paper, we apply Self-Attention into Attention-based MIL Pooling (SA-AbMILP), which combines the multi-level dependencies across image regions with the trainable operator of weighted average pooling. In contrast to Attention-based MIL Pooling (AbMILP), it covers not only the standard but also the presence-based and threshold-based assumptions of MIL. Self-Attention is detecting the relationship between instances, so it can embed into the instance feature vectors the information about the presence of similar instances or find that a combination of specific instances defines the bag as a positive. The experiments on five datasets (MNIST, two histological datasets of breast and colon cancer, microbiological dataset DIFaS, and retinal image screening) confirm that our method is on par or outperforms current state-of-the-art methodology based on the Wilcox pair test. We demonstrate that in the case of bigger datasets, it is advisable to use various kernels of the self-attention instead of the commonly used dot product. We also provide qualitative results to illustrate the reason for the improvements achieved by our method.

The experiments show that methods covering a wider range of MIL assumptions fit better for real-world problems. Therefore, in future work, we plan to introduce methods for more challenging MIL assumptions, e.g. collective assumption, and apply them to more complicated tasks, like digestive track assessment using the Nancy Histological Index. Moreover, we plan to introduce better interpretability, using Prototype Networks.

\section{Acknowledgments} 
The POIR.04.04.00-00-14DE/18-00 project is carried out within the Team-Net programme of the Foundation for Polish Science co-financed by the European Union under the European Regional Development Fund.

\bibliographystyle{ieee_fullname}
\bibliography{paper}

\end{document}